\documentclass[]{article}
\usepackage{naacl2001}
\usepackage{epsfig}
\title{Information Extraction Using the Structured Language Model}

\author{Ciprian Chelba and Milind Mahajan\\
        Microsoft Research \\
        Microsoft Corporation\\
        One Microsoft Way, Redmond, WA 98052\\
        {\tt \{chelba,milindm\}@microsoft.com}
}

\begin{document}
%\makeidpage
\maketitle
\begin{abstract}
The paper presents a data-driven approach to information
extraction (viewed as template filling) using the structured language model
(SLM) as a statistical parser. The task of template filling is cast as
constrained parsing using the SLM. The model is automatically trained
from a set of sentences annotated with frame/slot labels and
spans. Training proceeds in stages: first a constrained syntactic
parser is trained such that the parses on training data meet the
specified semantic spans, then the non-terminal labels are enriched to
contain semantic information and finally a constrained syntactic+semantic
parser is trained on the parse trees resulting from the previous
stage. Despite the small amount of training data used, the model is
shown to outperform the slot level accuracy of a simple semantic
grammar authored manually for the MiPad --- personal information
management --- task.
\end{abstract}

%----------------------------------------------------------------------------------------
\section{Introduction} \label{sec:intro} 

Information extraction from text can be characterized as template
filling~\cite{jurafsky:2000}: a given template or \emph{frame} contains a certain number
of \emph{slots} that need to be filled in with segments of text. Typically
not all the words in text are relevant to a particular frame. 
Assuming that the segments of text relevant to
filling in the slots are non-overlapping contiguous strings of words, one can
represent the semantic frame as a simple semantic parse tree for the
sentence to be processed. The tree has two levels: the root node is
tagged with the frame label and spans the entire sentence;
the leaf nodes are tagged with the slot labels and span the
strings of words relevant to the corresponding slot.
 
Consider the semantic parse $S$ for a sentence $W$ presented in Fig.~\ref{fig:s_parse}.
\begin{figure}[h]
(\verb+CalendarTask+ schedule meeting with (\verb+ByFullName*Person+ megan hokins) about (\verb+SubjectByWildCard*Subject+ internal lecture) at (\verb+PreciseTime*Time+ two thirty p.m.))
\caption{Sample sentence and semantic parse} \label{fig:s_parse}
\end{figure}
\verb+CalendarTask+ is the frame tag, spanning the
entire sentence; the remaining ones are slot tags with their
corresponding spans.

In the MiPad scenario~\cite{msft:MiPad} --- essentially a personal
information management (PIM) task --- there is a module that is able to convert the
information extracted according to the semantic parse into specific
actions. In this case the action is to schedule a calendar appointment.

We view the problem of information extraction as the recovery of the two-level
semantic parse $S$ for a given word sequence $W$. 

We propose a data driven approach to information extraction that uses
the structured language model (SLM)~\cite{chelba00} as an automatic
parser. The parser is \emph{constrained} to explore only parses that contain
pre-set constituents --- spanning a given word string and bearing a tag in a
given set of semantic tags. The constraints available during training
and test are different, the test case constraints being more relaxed as
explained in Section~\ref{sec:cparser_slm}.

The main advantage of the approach is that it doesn't require
any grammar authoring expertise. The approach is fully automatic once
the annotated training data is provided; it does assume that an
application schema --- i.e. frame and slot structure --- has been defined
but does not require semantic grammars that identify word-sequence to slot or
frame mapping. However, the process of converting the word sequence
coresponding to a slot into actionable canonical forms --- i.e. convert \emph{half past
two in the afternoon} into \emph{2:30 p.m.} --- may require
grammars. The design of the frames --- what information is relevant
for taking a certain action, what slot/frame tags are to be used,
see~\cite{msft:slu_parser} --- is a delicate task that we will not be
concerned with for the purposes of this paper.

The remainder of the paper is organized as follows:
Section~\ref{sec:slm} reviews the structured language model (SLM)
followed by 
Section~\ref{sec:train} which describes in detail the training
procedure and Section~\ref{sec:cparser_slm} which defines the operation of the
SLM as a constrained parser and presents the necessary modifications
to the model.
Section~\ref{sec:comparison} compares our approach to others in the
literature, in particular that of~\cite{miller00}.
Section~\ref{sec:experiments} presents the experiments we have
carried out. We conclude with Section~\ref{sec:conclusions}.

%----------------------------------------------------------------------------------------
\section{Structured Language Model} \label{sec:slm}

%\section{Structured Language Model} \label{sec:slm}
We proceed with a brief review of the structured language model (SLM);
an extensive presentation of the SLM can be found
in~\cite{chelba00}. The model assigns a probability $P(W,T)$ to every
sentence $W$ and its every possible binary parse $T$. The
terminals of $T$ are the words of $W$ with POStags, and the nodes of $T$ are
annotated with phrase headwords and non-terminal labels.
\begin{figure}[h]
  \begin{center}
    \epsfig{file=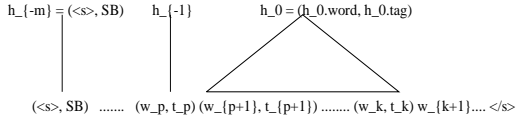,height=1.5cm,width=7cm}
  \end{center}
  \caption{A word-parse $k$-prefix} \label{fig:w_parse}
\end{figure}
 Let $W$ be a sentence of length $n$ words to which we have prepended
the sentence beginning marker \verb+<s>+ and appended the sentence end
marker \verb+</s>+ so that $w_0 = $\verb+<s>+ and $w_{n+1} = $\verb+</s>+.
Let $W_k = w_0 \ldots w_k$ be the word $k$-prefix of the sentence ---
the words from the begining of the sentence up to the current position
 $k$ --- and  \mbox{$W_k T_k$} the \emph{word-parse $k$-prefix}. Figure~\ref{fig:w_parse} shows a
word-parse $k$-prefix; \verb|h_0 .. h_{-m}| are the \emph{exposed
 heads}, each head being a pair (headword, non-terminal label), or
(word,  POStag) in the case of a root-only tree. The exposed heads at
a given position $k$ in the input sentence are a function of the
word-parse $k$-prefix.

\subsection{Probabilistic Model} \label{ssec:prob_model}

\begin{figure}[h]
  \begin{center} 
    \epsfig{file=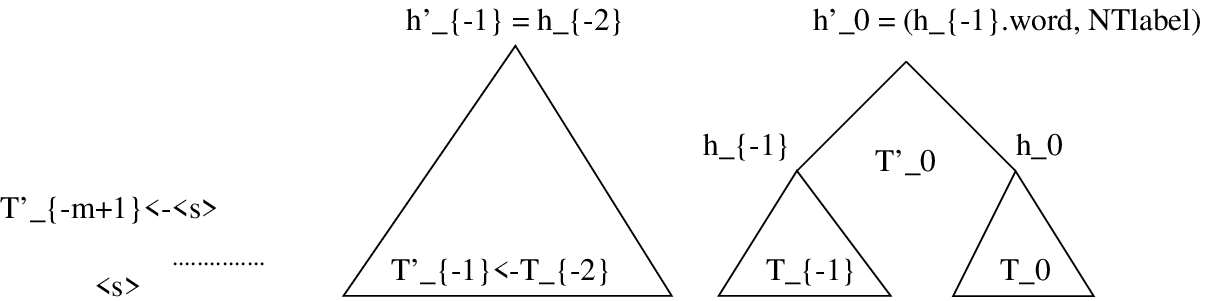,height=1.7cm,width=7cm}
  \end{center}
  \caption{Result of adjoin-left under NTlabel} \label{fig:after_a_l}
%\end{figure}
%\begin{figure}
  \begin{center} 
    \epsfig{file=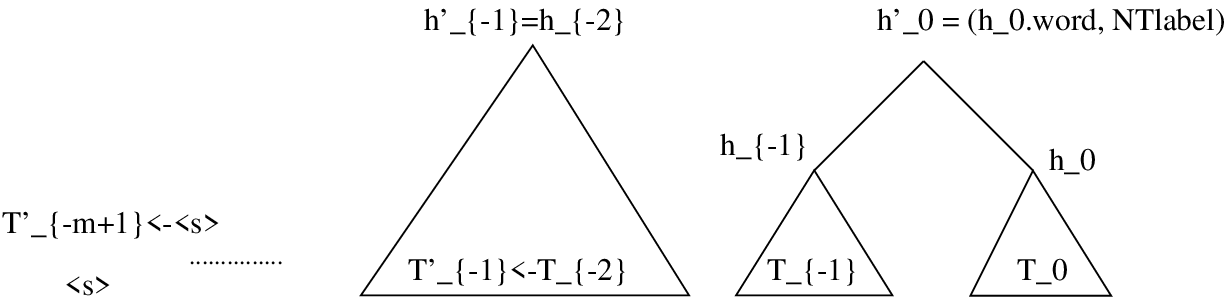,height=1.7cm,width=7cm}
  \end{center}
  \caption{Result of adjoin-right under NTlabel} \label{fig:after_a_r}
\end{figure}

 The joint probability $P(W,T)$ of a word sequence $W$ and a complete parse
$T$ can be broken into:
\begin{eqnarray}
\lefteqn{P(W,T)= } \nonumber\\%\label{eq:model}\\
& \prod_{k=1}^{n+1}[P(w_k/W_{k-1}T_{k-1}) \cdot P(t_k/W_{k-1}T_{k-1},w_k) \cdot \nonumber\\
& \prod_{i=1}^{N_k}P(p_i^k/W_{k-1}T_{k-1},w_k,t_k,p_1^k\ldots
p_{i-1}^k)] \nonumber
\end{eqnarray}
where:
\begin{itemize}
\item $W_{k-1} T_{k-1}$ is the word-parse $(k-1)$-prefix
\item $w_k$ is the word predicted by WORD-PREDICTOR
\item $t_k$ is the tag assigned to $w_k$ by the TAGGER
\item $N_k - 1$ is the number of operations the PARSER executes at 
sentence position $k$ before passing control to the  WORD-PREDICTOR
(the $N_k$-th operation at position k is the \verb+null+ transition);
$N_k$ is a function of $T$
\item $p_i^k$ denotes the $i$-th PARSER operation carried out at
position k in the word string; the operations performed by the
PARSER are illustrated in
Figures~\ref{fig:after_a_l}-\ref{fig:after_a_r} and they ensure that
all possible binary branching parses with all possible headword and
non-terminal label assignments for the $w_1 \ldots w_k$ word
sequence can be generated. The $p_1^k \ldots p_{N_k}^k$ sequence of PARSER
operations at position $k$ grows the word-parse $(k-1)$-prefix into a
word-parse $k$-prefix.
\end{itemize}
%\begin{figure}
%  \begin{center} 
%    \epsfig{file=before.eps,height=1.3cm,width=7cm}
%  \end{center}
%  \caption{Before an adjoin operation} \label{fig:before}
%\end{figure}
 
Our model is based on three probabilities, each estimated using deleted
interpolation and parameterized (approximated) as follows:
\begin{eqnarray}
  P(w_k/W_{k-1} T_{k-1}) 	& \doteq & P(w_k/h_0, h_{-1})\nonumber\label{eq:1}\\
  P(t_k/w_k,W_{k-1} T_{k-1}) 	& \doteq & P(t_k/w_k, h_0, h_{-1})\nonumber\label{eq:2}\\
  P(p_i^k/W_k T_k) & \doteq & P(p_i^k/h_0, h_{-1})\nonumber\label{eq:3}
\end{eqnarray}%
 It is worth noting that if the binary branching structure
developed by the parser were always right-branching and we mapped the
POStag and non-terminal label vocabularies to a single type then our
model would be equivalent to a trigram language model.
 Since the number of parses  for a given word prefix $W_{k}$ grows
exponentially with $k$, $|\{T_{k}\}| \sim O(2^k)$, the state space of
our model is huge even for relatively short sentences, so we had to use
a search strategy that prunes it. Our choice was a synchronous
multi-stack search algorithm which is very similar to a beam search.

The \emph{language model} probability assignment for the word at position $k+1$ in the input
sentence is made using:
\begin{eqnarray}
P(w_{k+1}/W_{k}) & =
& \sum_{T_{k}\in
  S_{k}}P(w_{k+1}/W_{k}T_{k})\cdot\rho(W_{k}T_{k}),\nonumber \\ 
\rho(W_{k}T_{k}) & = & P(W_{k}T_{k})/\sum_{T_{k} \in S_{k}}P(W_{k}T_{k})\label{eq:ppl1}
\end{eqnarray}
which ensures a proper probability over strings $W^*$, where $S_{k}$ is
the set of all parses present in our stacks at the current stage $k$.

\subsection{Model Parameter Estimation} \label{ssec:model_estimation}

Each model component --- WORD-PREDICTOR, TAGGER, PARSER ---
is initialized from a set of parsed sentences after undergoing
headword percolation and binarization. 
Separately for each model component we:\\
\begin{itemize}
\item gather counts from ``main'' data --- about 90\% of the
training data
\item estimate the interpolation coefficients on counts gathered from
``check'' data --- the remaining 10\% of the training data.
\end{itemize}

An N-best EM~\cite{em77} variant is then employed to jointly
re-estimate the model parameters such that the likelihood of the
training data under our model is increased.

%----------------------------------------------------------------------------------------
\section{Training Procedure} \label{sec:train}

This section describes the training procedure for the SLM when
applied to information extraction and introduces the modifications that need to
be made to the SLM operation.\\

The training of the model proceeds in four stages:
\begin{enumerate}
\item initialize the SLM as a syntactic parser for the domain we are
interested in. A general purpose parser (such as NLPwin~\cite{msft:nlpwin})
can be used to generate a syntactic treebank from which the SLM
parameters can be initialized. Another possibility for initializing
the SLM is to use a treebank for out-of-domain data (such as the UPenn
Treebank~\cite{Upenn}) --- see Section~\ref{sec:experiments_upenn}.

\item train the SLM as a \emph{matched constrained parser}. 
At this step the parser is going to propose a set of N syntactic
binary parses for a given word string (N-best parsing), all matching
the constituent boundaries specified by the semantic parse: a parse
$T$ is said to \emph{match} the semantic parse $S$, denoted $T \ni S$,
if and only if the set of un-labeled constituents that define $S$ is
included in the set of constituents that define $T$.

At this time only the constituent span information in $S$ is taken
into account.

\item enrich the non-terminal and pre-terminal labels of the resulting
parses with the semantic tags (frame and slot) present in the semantic
parse, thus expanding the vocabulary of non-terminal and pre-terminal
tags used by the syntactic parser to include semantic information
alongside the usual syntactic tags.

\item train the SLM as a \emph{L(abel)-matched constrained parser}
to explore only the semantic parses for the training data. This time the semantic
constituent labels are taken into account too, which means that a
parse $P$ --- containing both syntactic and semantic information ---
is said to \emph{L(abeled)-match} $S$ if and only if the set of
labeled semantic constituents that defines $S$ is identical to the set of
semantic constituents that defines $P$. If we let $SEM(P)$ denote the
function that maps a tree $P$ containing both syntactic and semantic
information to the tree containing only semantic information, referred
to as the \emph{semantic projection} of $P$, then all
the parses  $P_i, \forall i < N$, proposed by the SLM for a given
sentence $W$, L-match $S$ and thus satisfy $SEM(P_i) = S, \forall i < N$.

The semantic tree $S$ has a two level structure
so the above requirement can be satisfied only if the parses $SEM(P)$
proposed by the SLM are also on two levels, frame and slot level
respectively. We have incorporated this constraint into the operation
of the SLM --- see Section~\ref{subsec:layering}.
\end{enumerate}

The model thus trained is then used to parse test sentences and
recover the semantic parse using $S = SEM(\arg \max_{P_i} P(P_i,W))$. 
In principle, one should sum over all the parses $P$
that yield the same semantic parse $S$ and then choose 
$S = \arg \max_{S} \sum_{P_i s.t.\ SEM(P_i) = S} P(P_i,W)$. 

A few iterations of the N-best EM variant 
--- see Section~\ref{sec:slm} --- were run at each of the
second and fourth step in the training procedure. The constrained
parser operation makes this an EM variant where the hidden space ---
the possible parse trees for a given sentence --- is a
priori limited by the semantic constraints to a subset of the hidden space
of the unrestricted model. At test time we
wish to recover the most likely subset of the hidden space
consistent with the constraints imposed on the sentence.

To be more specific, during the second training stage, the E-step of
the reestimation procedure will only explore
syntactic trees (hidden events) that \emph{match} the semantic parse; the fourth
stage E-steps will consider hidden events that are constrained even further to
\emph{L-match} the semantic parse. We have no proof that this
procedure should lead to better performance in terms of slot/frame accuracy
but intuitively one expects it to place more and more probability mass
on the desirable trees --- that is, the trees that are consistent with
the semantic annotation. This is confirmed experimentally by the fact
that the likelihood of the training word sequence (observable)
--- calculated by Eq.~(\ref{eq:ppl1}) where the sum runs over the parse
trees that \emph{match/L-match} the semantic constraints --- does
increase\footnote{Equivalent to a decrease in perplexity} at every
training step, as presented in Section~\ref{sec:experiments},
Table~\ref{table:likelihood}. However, the increase in likelihood is not
always correlated with a decrease in error rate on the training
data, see Tables~\ref{table:accuracies}~and~\ref{table:accuracies_UPenn} in
Section~\ref{sec:experiments}.

%----------------------------------------------------------------------------------------
\section{Constrained Parsing Using the Structured Language Model} \label{sec:cparser_slm}

We now detail the constrained operation of the SLM --- \emph{matched} and
\emph{L-matched} parsing --- used at the second and fourth steps of the
training procedure described in the previous section.

A semantic parse $S$ for a given sentence $W$ consists of a set of
constituent boundaries along with semantic tags. When parsing the
sentence using the standard formulation of the SLM, one obtains binary
parses that are not guaranteed to match the semantic parse $S$,
i.e. the constituent proposed by the SLM may cross semantic
constituent boundaries; for the constituents matching the semantic
constituent boundaries, the labels proposed may not be the desired ones.

To fix terminology, we define a constrained constituent --- or simply
a constraint --- $c$ to be a span together with a 
\emph{set}\footnote{The set of allowable tags must contain at least one
element} of allowable tags for the span: $c = <l,r,Q>$ where $l$ is
the left boundary of the constraint, $r$ is the right boundary of the
constraint and $Q$ is the set of allowable non-terminal tags for the
constraint.% taking value in a vocabulary $\mathcal{V}_N$.

A semantic parse can be viewed as a set of constraints; for each
constraint the set of allowable non-terminal tags $Q$ contains a single
element, respectively the semantic tag for each constituent. An
additional fact to be kept in mind is that the semantic parse tree
consists of exactly two levels: the frame level (root semantic tag)
and the slot level (leaf semantic tags).

During training, we wish to constrain the SLM operation such that it
considers only parses that match the constraints $c_i, i=1 \ldots C$
as it proceeds left to right through a given sentence $W$. In light of
the training procedure sketched in the introduction, we consider two
flavors of constrained parsing, one in which we only generate parses
that \emph{match} the constraint boundaries and another in which we
also enforce that the proposed tag for every matching constituent is
among the constrained set of non-terminal tags $c_i.Q$ ---
\emph{L(abeled)-match} constrained parsing.

The only constraints available for the test sentences are: 
\begin{itemize}
\item the semantic tag of the root node --- which spans the entire
sentence --- must be in the set of frame
tags. If it were a test
sentence the example in Figure~\ref{fig:s_parse}
would have the following semantic parse (constraints):
(\verb+{CalendarTask,ContactsTask,MailTask}+ schedule 
meeting with megan hokins about internal lecture to two thirty p.m.)

\item the semantic projection of the trees proposed by the SLM must
have exactly two levels; this constraint is built in the operation of
the L-match parser.
\end{itemize}

The next section will describe the constrained parsing algorithm. 
Section~\ref{subsec:layering} will describe further changes
that the algorithm uses to produce only parses $P$ whose semantic
projection $SEM(P)$ has
exactly two levels, frame (root) and slot (leaf) level, respectively
--- only in the L-match case.
We conclude with Section~\ref{subsec:pruning} explaining how the
constrained parsing algorithm interacts with the pruning of the SLM
search space for the most likely parse.

\subsection{Match and L-match SLM Parsing} \label{subsec:match_parsing}

The trees produced by the SLM are binary trees. The tags annotating
the nodes of the tree are purely syntactic --- during the second
training stage --- or syntactic+semantic --- during the last training
stage or at test time. It can be proved that satisfying the following two
conditions at each position $k$ in the input sentence ensures that all
the binary trees generated by the SLM parsing algorithm match
the pre-set constraints $c_i, i=1 \ldots C$ as it proceeds
left to right through the input sentence $W = w_0 \ldots w_{n+1}$.
\begin{itemize}
\item for a given word-parse $k$-prefix \mbox{$W_k T_k$} (see
Section~\ref{sec:slm}) accept an \verb+adjoin+ transition if and only
if:
\begin{enumerate}
\item	the resulting constituent does not violate\footnote{A constraint is
	violated by a constituent if the span of the constituent crosses the
	span of the  constraint.} any of the constraints $c_i, i=1 \ldots C$
\item	\emph{L-match} parsing only: if the semantic projection of the
non-terminal tag $SEM(NTtag)$ proposed by the adjoin operation is
non-void then the newly created constituent must L-match an existing
constraint, $\exists\ c_i\ s.t.\ SEM(NTtag) \in c_i.Q$.
\end{enumerate}

\item for a given word-parse $k$-prefix \mbox{$W_k T_k$} (see
Section~\ref{sec:slm}) accept the \verb+null+ transition if and only
if all the constraints $c_i$ whose right boundary is equal to the
current word index $k$, $c_i.r = k$, have been matched. If these
constraints remain un-matched they will be broken at a later time
during the process of completing the parse for the current sentence
$W$: there will be an adjoin operation involving a constituent to the
right of the current position that will break all the constraints
ending at the current position $k$. 
\end{itemize}

\subsection{Semantic Tag Layering} \label{subsec:layering}

The two-layer structure of the semantic trees need not be 
enforced during training, simply L-matching the semantic constraints
will implicitly satisfy this constraint. 
As explained above, for test sentences we can only specify the frame
level constraint, leaving open the possibility of generating a tree
whose semantic projection would contain more than two levels --- nested
slot level constituents. In order to avoid this, each tree in a given
word-parse has two bits that describe whether the tree already contains a
constituent whose semantic projection is a frame/slot level tag,
respectively. An adjoin operation proposing a tag that violates the
correct layering of frame/slot level tags can now be detected and discarded.

\subsection{Interaction with Pruning} \label{subsec:pruning}

In the absence of pruning the search for the most likely parse
satisfying the constraints for a given sentence
becomes computationally intractable\footnote{It is assumed that the constraints for a
given sentence are consistent, namely there exists at least one parse
that meets all of them.}. In practice, we are forced to use pruning
techniques in order to limit the size of the search space. However, it is
possible that during the left to right traversal of the
sentence, the pruning scheme will keep alive only parses
whose continuation cannot meet constraints that we have not
encountered yet and no complete parse for the current sentence can be
returned. In such cases, we back-off to unconstrained parsing ---
regular SLM usage. In our experiments, we noticed that this was
necessary for very few training sentences (1 out of 2,239) and
relatively few test sentences (31 out of 1,101).

%----------------------------------------------------------------------------------------
\section{Comparison with Previous Work} \label{sec:comparison}

The use of a syntactic parser augmented with semantic tags for
information information from text is not a novel idea. The basic
approach we described is very similar to the one presented
in~\cite{miller00} however there are a few major differences:
\begin{itemize}
\item in our approach the augmentation of the syntactic tags with semantic tags is
straightforward due to the fact that the semantic constituents are
matched exactly\footnote{This is a consequence of the fact that the
SLM generates binary trees}. The approach in~\cite{miller00} needs to insert
additional nodes in the syntactic tree to account for the semantic
constituents that do not have a corresponding syntactic one. We
believe our approach ensures tighter coupling between the syntactic and
the semantic information in the final augmented trees.
\item our constraint definition allows for a set of semantic tags to
be matched on a given span.
\item the two-level layering constraint on semantic trees is a
structural constraint that is embedded in the operation of the SLM and
thus can be guaranteed on test sentences. 
\end{itemize}

The semantic annotation required by our task is much simpler than that
employed by~\cite{miller00}. One possibly beneficial extension of our work
suggested by~\cite{miller00} would be to add semantic tags describing
relations between entities (slots), in which case the semantic
constraints would not be structured strictly on the two levels used in
the current approach, respectively frame and slot level. However, this
would complicate the task of data annotation making it more expensive. 

The same constrained EM variant employed for reestimating the model parameters has been used
by~\cite{pereira_schabes92} for training a purely syntactic parser
showing increase in likelihood but no improvement in parsing accuracy.

%----------------------------------------------------------------------------------------
\section{Experiments} \label{sec:experiments}

We have evaluated the model on manually annotated data for the
MiPad~\cite{msft:MiPad} task. We have used 2,239 sentences (27,119
words) for training and 1,101 sentences (8,652 words) for test. There
were 2,239/5,431 semantic frames/slots in the training data and 1,101/1,698 in
the test data, respectively.

The word vocabulary size was 1,035, closed over the test data. The slot
and frame vocabulary sizes were 79 and 3, respectively. The
pre-terminal (POStag) vocabulary sizes were 64 and 144 for training
steps 2 and 4 (see Section~\ref{sec:train}), respectively;
the non-terminal (NTtag) vocabulary sizes were 61 and 540 for training
steps 2 and 4 (see Section~\ref{sec:train}), respectively. We have
used the NLPwin~\cite{msft:nlpwin} parser to obtain the MiPad syntactic
treebank needed for initializing the SLM at training step 1.

\begin{table}[h]
  \begin{center}
    \begin{tabular}{|l|r|r|r|} \hline
      \multicolumn{2}{|c|}{Training}&
      \multicolumn{2}{c|}{Perplexity}\\
      Stage 		& It 	& Training Set 	& {Test set}	\\\hline
      2 (matched) 	& 0 	& 9.27		& 34.81		\\ 
      2 (matched)	& 1 	& 5.81	 	& 31.25	 	\\ 
      2 (matched)	& 2 	& 5.51		& 31.41	 	\\ \hline
      4 (L-matched)	& 0 	& 4.71 		& 24.39 	\\ 
      4 (L-matched)	& 1 	& 4.61		& 24.73		\\ 
      4 (L-matched)	& 2 	& 4.56		& 24.88 	\\ \hline
    \end{tabular}
    \caption{Likelihood Evolution during Training}
    \label{table:likelihood}
  \end{center}
\end{table}
% table showing the evolution of the trainign/test data likelihood

% 3. evolution of likelihood on train/test during the two EM steps.
Although not guaranteed theoretically, the N-best EM variant used for the SLM
parameter reestimation increases the likelihood of the training data
with each iteration when the parser is run in both matched
(training step 2) and L-matched (training step 4) constrained modes.
Table~\ref{table:likelihood} shows the evolution of the
training and test data perplexities (calculated using the probability
assignment in Eq.~\ref{eq:ppl1}) during the constrained training steps 2 and 4.

The  training data perplexity decreases monotonically
during both training steps whereas
the test data perplexity doesn't decrease monotonically in either
case. We attribute this discrepancy between the evolution of the
likelihood on the training and test corpora to the different
constrained settings for the SLM.

The most important performance measure is the slot/frame error rate.
To measure it, we use manually created parses which consist of frame-level labels and
slot-level labels and spans as reference. A frame-level error is
caused by a frame label of the hypothesis parse which is different
from the frame label of the reference. In order to calculate the
slot-level errors, we create a set of slot label and slot span pairs for
the reference and hypothesis parse, respectively. The number of slot errors is then
the minimum edit distance between these 2 sets using the substitution,
insertion and deletion operations on the elements of the set. 

\begin{table*}
  \begin{center}
    \begin{tabular}{|l|l|r|r|r|r|r|r|} \hline
      \multicolumn{2}{|c|}{Training It} & \multicolumn{6}{c|}{Error Rate (\%)}\\
      \multicolumn{2}{|l|}{} & \multicolumn{2}{c|}{Training} & \multicolumn{2}{c|}{Test} & \multicolumn{2}{c|}{Test-L1}\\
      Stage 2 	& Stage 4 	& Slot 	& Frame & Slot 	& Frame & Slot 	& Frame 	\\\hline
\multicolumn{2}{|c|}{Baseline}
			& 43.41	& 7.20	& 57.36 & 14.90 & 57.30	& 6.90  \\ \hline
      0 	& 0 	& 9.78	& 1.65	& \underline{37.87} & \underline{21.62} & 37.46	& 0.64 	\\ 
      0 	& 1 	& 10.36 & 1.20 	& 39.16	& 21.80	& 38.28	& 0.64	\\ 
      0 	& 2 	& 9.42 	& 1.05 	& 39.75	& 22.25	& 38.63	& 0.82	\\\hline  
      2 	& 0 	& 8.92 	& 1.25 	& 38.04	& 22.07	& 37.81	& 0.91 	\\ 
      2 	& 1 	& 9.01 	& 0.95 	& 37.51	& 21.89	& 37.28	& 0.91	\\ 
      2 	& 2 	& 9.47 	& 0.90 	& 38.99	& 21.89	& 38.57	& 0.82	\\\hline
    \end{tabular}
    \caption{Training and Test Data Slot/Frame Error Rates}
    \label{table:accuracies}
  \end{center}
%\end{table*}
% table showing the evolution of the trainign/test data error rates
%\begin{table*}
  \begin{center}
    \begin{tabular}{|l|l|r|r|r|r|r|r|} \hline
      \multicolumn{2}{|c|}{Training It} & \multicolumn{6}{c|}{Error Rate (\%)}\\
      \multicolumn{2}{|l|}{} & \multicolumn{2}{c|}{Training} & \multicolumn{2}{c|}{Test} & \multicolumn{2}{c|}{Test-L1}\\
      Stage 2 	& Stage 4 	& Slot 	& Frame & Slot 	& Frame & Slot 	& Frame 	\\\hline
\multicolumn{2}{|c|}{Baseline}
			& 43.41	& 7.20	& 57.36 & 14.90 & 57.30	& 6.90  \\ \hline
      0, MiPad/NLPwin	& 0 	& 9.78	& 1.65	& 37.87 & 21.62	& 37.46	& 0.64 	\\ \hline
      1, UPenn Trbnk 	& 0 	& 8.44	& 2.10	& \underline{36.93} & \underline{16.08} & 36.34	& 0.91 	\\
      1, UPenn Trbnk	& 1 	& 7.82 	& 1.70 	& 36.98	& 16.80	& 36.22	& 0.82	\\
      1, UPenn Trbnk	& 2 	& 7.69 	& 1.50 	& 36.98	& 16.80	& 36.22	& 1.00	\\ \hline
    \end{tabular}
    \caption{Training and Test Data Slot/Frame Error Rates, UPenn
Treebank initial statistics}
    \label{table:accuracies_UPenn}
  \end{center}
%\end{table*}
% table showing the evolution of the trainign/test data error rates,
% Upenn stats
%\begin{table*}
  \begin{center}
    \begin{tabular}{|l|l|l|r|r|r|r|r|r|} \hline
      Training & \multicolumn{2}{|c|}{Training It} & \multicolumn{6}{c|}{Error Rate (\%)}\\
      Corpus   & \multicolumn{2}{|l|}{} & \multicolumn{2}{c|}{Training} & \multicolumn{2}{c|}{Test} & \multicolumn{2}{c|}{Test-L1}\\
      Size     & Stage 2 	& Stage 4 	& Slot 	& Frame & Slot 	& Frame & Slot 	& Frame 	\\\hline
      \multicolumn{3}{|c|}{Baseline}
			 		& 43.41	& 7.20	& 57.36 & 14.90 & 57.30	& 6.90  \\ \hline
      all      & 1, UPenn Trbnk & 0 	& 8.44	& 2.10	& \underline{36.93} & \underline{16.08} & 36.34	& 0.91 	\\ 
      1/2 all  & 1, UPenn Trbnk	& 0 	& --- 	& --- 	& 43.76	& 18.44	& 43.40	& 0.45	\\ 
      1/4 all  & 1, UPenn Trbnk	& 0 	& --- 	& --- 	& 49.47	& 22.98	& 49.53	& 1.82	\\ \hline
    \end{tabular}
    \caption{Performance Degradation with Training Data Size}
    \label{table:accuracies_UPenn_tr_size}
  \end{center}
\end{table*}

Table~\ref{table:accuracies} shows the error rate on training and test
data at different stages during training. The last column of test data
results (Test-L1) shows the results obtained by assuming that the user
has specified the identity of the frame --- and thus the frame level
constraint contains only the correct semantic tag. This is a plausible
scenario if the user has the possibility to choose the frame using a
different input modality such as a stylus. The error rates on the training
data were calculated by running the model with the same constraint as on the test
data --- constraining the set of allowable tags at the frame level. This
could be seen as an upper bound on the performance of the model (since
the model parameters were estimated on the same data). 

Our model significantly outperforms the baseline model --- a simple
semantic context free grammar authored manually for the MiPad task ---
in terms of slot error rate (about 35\% relative reduction in slot
error rate) but it is outperformed by the latter in
terms of frame error rate. When running the models from training step 2 on test data one cannot add any
constraints; only frame level constraints can be used when evaluating
the models from training step 4 on test data.
N-best reestimation at either
training stage (2 or 4) doesn't improve the accuracy of the system,
although the results obtained by intializing the model using the
reestimated stage 2 model --- iteration 2-\{0,1,2\} models tend to be
slightly better than their 0-\{0,1,2\} counterparts. Constraining the frame
level tag to have the correct value doesn't significantly reduce the slot error
rate in either approach, as can be seen from the Test-L1 column of results\footnote{The
frame error rate in this column should be 0; in practice this doesn't
happen because some test sentences could not be L-match parsed
using the pruning strategy employed by the SLM, see Section~\ref{subsec:pruning}}. 

\subsection{Out-of-domain Initial Statistics} \label{sec:experiments_upenn}

Recent results~\cite{chelba01} on the portability of syntactic
structure within the SLM framework show that it is possible to
initialize the SLM parameters from a treebank for out-of-domain text and
maintain the same language modeling performance. We have repeated the
experiment in the context of information extraction. 

Similar to the approach in~\cite{miller00} we
initialized the SLM statistics from the UPenn Treebank parse trees
(about 1Mwds of training data) at the first training stage, see
Section~\ref{sec:train}. The remaining part of the training procedure
was the same as in the previous set of experiments.

The word, slot and frame vocabulary were the same as in the previous set of experiments. The
pre-terminal (POStag) vocabulary sizes were 40 and 204 for training
steps 2 and 4 (see Section~\ref{sec:train}), respectively;
the non-terminal (NTtag) vocabulary sizes were 52 and 434 for training
steps 2 and 4 (see Section~\ref{sec:train}), respectively.

The results are presented in Table~\ref{table:accuracies_UPenn}, showing improved performance over
the model initialized from in-domain parse trees. The frame accuracy
increases substantially, almost matching that of the baseline model,
while the slot accuracy is just slightly increased. We attribute the
improved performance of the model initialized from the UPenn Treebank
to the fact that the model explores a more diverse set of trees for a given
sentence than the model initialized from the MiPad automatic treebank
generated using the NLPwin parser.

\subsection{Impact of Training Data Size on Performance} \label{sec:experiments_upenn_tr_size}

We have also evaluated the impact of the training data size on the
model performance. The results are presented in
Table~\ref{table:accuracies_UPenn_tr_size}, showing a strong
dependence of both the slot and frame error rates on the amount of
training data used. This, together with the high accuracy of the model on
training data (see Table~\ref{table:accuracies_UPenn}), suggests
that we are far from saturation in performance and that more training
data is very likely to improve the model performance substantially.

\subsection{Error Trends} \label{sec:experiments_error}

As a summary error analysis, we have investigated the correlation
between the semantic frame/slot error rate and the number of semantic
slots in a sentence. We have binned the sentences in the test set
according to the number of slots in the manual annotation and
evaluated the frame/slot error rate in each bin. The results are shown
in Table~\ref{table:error_sent}. 

The frame/slot accuracy increases with the number of slots per
sentence --- except for the 5+ bin where the frame error rate
increases --- showing that slot co-ocurence statistics improve
performance; sentences containing more semantic slots tend to be less
ambiguous from an information extraction point of view.
\begin{table}[h]
  \begin{center}
    \begin{tabular}{|l|r|r|r|} \hline
      		 	& \multicolumn{2}{l|}{Error Rate (\%)} & \\
       	No. slots/sent	& Slot  & Frame  	& No. Sent \\\hline
	1		& 43.97	& 18.01	& 755	\\
	2		& 39.23	& 16.27	& 209	\\
	3		& 26.44	&  5.17	&  58	\\
	4		& 26.50	&  4.00	&  50	\\
	5+		& 21.19	&  6.90	&  29	\\\hline
    \end{tabular}
    \caption{Frame/Slot Error Rate versus Slot Density}
    \label{table:error_sent}
  \end{center}
\end{table}

%----------------------------------------------------------------------------------------
\section{Conclusions and Future Directions} \label{sec:conclusions}

We have presented a data-driven approach to information extraction
that, despite the small amount of training data used, is
shown to outperform the slot level accuracy of a simple semantic
grammar authored manually for the MiPad --- personal information
management --- task. 

The performance of the baseline model could
be improved with more authoring effort, although this is expensive.

The big difference in performance between training and test
and the fact that we are using so little training data, makes
improvements by using more training data very likely, although this may be
expensive. A framework which utilizes the vast amounts of text data collected once such a
system is deployed would be desirable. Statistical modeling techniques that make more
effective use of the training data should be used in the SLM, maximum
entropy~\cite{berger:max_ent} being a good candidate.

As for using the SLM as the language understanding component of a
speech driven application, such as MiPad, it would be interesting to
evaluate the impact of incorporating the semantic constraints on the
word-level accuracy of the system. Another possible research direction
is to modify the framework such that it finds the most likely semantic
parse given the acoustics --- thus treating the word sequence as a
hidden variable.
%----------------------------------------------------------------------------------------
\bibliographystyle{acl}
\bibliography{submission}

\begin{thebibliography}{}

\bibitem[\protect\citename{Berger \bgroup et al.\egroup }1996]{berger:max_ent}
A.~L. Berger, S.~A.~Della Pietra, and V.~J.~Della Pietra.
\newblock 1996.
\newblock A maximum entropy approach to natural language processing.
\newblock {\em Computational Linguistics}, 22(1):39--72, March.

\bibitem[\protect\citename{Chelba and Jelinek}2000]{chelba00}
Ciprian Chelba and Frederick Jelinek.
\newblock 2000.
\newblock Structured language modeling.
\newblock {\em Computer Speech and Language}, 14(4):283--332, October.

\bibitem[\protect\citename{Chelba}2001]{chelba01}
Ciprian Chelba.
\newblock 2001.
\newblock Portability of syntactic structure for language modeling.
\newblock In {\em Proceedings of ICASSP}, page to appear. Salt Lake City, Utah.

\bibitem[\protect\citename{Dempster \bgroup et al.\egroup }1977]{em77}
A.~P. Dempster, N.~M. Laird, and D.~B. Rubin.
\newblock 1977.
\newblock Maximum likelihood from incomplete data via the {EM} algorithm.
\newblock In {\em Journal of the Royal Statistical Society}, volume~39 of {\em
  B}, pages 1--38.

\bibitem[\protect\citename{Heidorn}1999]{msft:nlpwin}
George Heidorn.
\newblock 1999.
\newblock Intelligent writing assistance.
\newblock In R.~Dale, H.~Moisl, and H.~Somers, editors, {\em Handbook of
  Natural Language Processing}. Marcel Dekker, New York.

\bibitem[\protect\citename{Huang \bgroup et al.\egroup }2000]{msft:MiPad}
X.~Huang, A.~Acero, C.~Chelba, L.~Deng, D.~Duchene, J.~Goodman, H.~Hon,
  D.~Jacoby, L.~Jiang, R.~Loynd, M.~Mahajan, P.~Mau, S.~Meredith, S.~Mughal,
  S.~Neto, M.~Plumpe, K.~Wang, and Y.~Wang.
\newblock 2000.
\newblock {MiPad}: A next generation {PDA} prototype.
\newblock In {\em ICSLP'00, Proceedings}, Beijing, China.

\bibitem[\protect\citename{Jurafsky and Martin}2000]{jurafsky:2000}
Daniel Jurafsky and James~H. Martin, 2000.
\newblock {\em An Introduction to Natural Language Processing, Computational
  Linguistics, and Speech Recognition}, pages 577--583.
\newblock Prentice Hall.

\bibitem[\protect\citename{Marcus \bgroup et al.\egroup }1993]{Upenn}
M.~Marcus, B.~Santorini, and M.~Marcinkiewicz.
\newblock 1993.
\newblock Building a large annotated corpus of {English}: the {Penn}
  {Treebank}.
\newblock {\em Computational Linguistics}, 19(2):313--330.

\bibitem[\protect\citename{Miller \bgroup et al.\egroup }2000]{miller00}
Scott Miller, Heidi Fox, Lance Ramshaw, and Ralph Weischedel.
\newblock 2000.
\newblock A novel use of statistical parsing to extract information from text.
\newblock In {\em Proceedings of ANLP-NAACL}, pages 226--233. Seattle,
  Washington.

\bibitem[\protect\citename{Pereira and Schabes}1992]{pereira_schabes92}
Fernando Pereira and Yves Schabes.
\newblock 1992.
\newblock Inside-outside reestimation from partially bracketed corpora.
\newblock {\em ACL}, 30:128--135.

\bibitem[\protect\citename{Wang}1999]{msft:slu_parser}
Y.-Y. Wang.
\newblock 1999.
\newblock A robust parser for spoken language understanding.
\newblock In {\em Eurospeech'99 Proceedings}, Budapest, Hungary.

\end{thebibliography}

\end{document}